\documentclass{article}
\usepackage{smc2026}
\usepackage[caption=false, font=footnotesize]{subfig}
\usepackage{paralist}
\usepackage[figure,table]{hypcap}
\usepackage{enumitem}

\usepackage[english]{babel}

\def\papertitle{Humanoid Musical Robots as Experimental Interfaces for Music-Evoked Emotion}

\author[1]{\mbox{\firstname{Vincent K.M.}\lastname{Cheung}\email{cheung@csl.sony.co.jp}}}
\author[2]{\mbox{\firstname{Jia-Yeu}\lastname{Lin}\email{erin@aoni.waseda.jp}}}

\affil[1]{\institution{Sony Computer Science Laboratories}\city{Tokyo}\country{Japan}\affiliationtype{Company}}
\affil[2]{\department{Faculty of Science and Engineering}\institution{Waseda University}\city{Tokyo}\country{Japan}\affiliationtype{University}}

\completesetup

\title{\papertitle}
\begin{document}
	\capstartfalse
	\maketitle
	\capstarttrue

\begin{abstract}


Advances in technology have led to increasingly sophisticated musical humanoid robots. However, their use has largely been limited to performance and related research in human–robot interaction. In this position paper, we propose a novel perspective: musical humanoid robots as experimental interfaces for investigating music-evoked emotions. We argue that current research is constrained by paradigms relying on pre-recorded auditory stimuli, which fail to capture the multimodal, embodied, and interactive nature of real-world musical experience. Building on existing theories of music cognition and emotion, we identify mechanisms that require controlled manipulation of both acoustic and non-acoustic variables. We show that humanoid robots are well-suited as they enable parametric control of performance variables, reproducibility across trials, and the decoupling and recombination of auditory, visual, and interactive components. We illustrate the technical feasibility of this perspective through a case study of the WAseda Saxophonist Robot 5 (WAS-5), demonstrating reproducible control of acoustic and interaction variables that are prerequisites for future music-emotion experiments. Our work positions musical humanoid robots as a methodological platform that enables future controlled investigations of music-evoked emotions.
\end{abstract}

\section{Introduction}\label{sec:introduction}

Music is a powerful medium of communication that modulates human emotions \cite{eerola2018music, mehr2019universality, koelsch2014brain}. A central goal in music cognition research is to understand how musical structures are perceived, integrated, and give rise to affective responses \cite{koelsch2011toward, rohrmeier2012predictive}.

Despite substantial progress over the past decades, much of this knowledge has been derived from laboratory paradigms relying on preselected, pre-recorded musical stimuli \cite{eerola2012review}. Such approaches implicitly assume that auditory information alone sufficiently captures the expressive content of musical performance and the mechanisms underlying music-evoked emotions. However, this assumption overlooks the embodied, multimodal, and interactive nature of real-world musical experience, where visual, motor, and social cues contribute to how music is perceived and felt \cite{koelsch2014brain, eerola2018music, juslin2008emotional}.

In parallel, advances in sensors, actuators, and computation have led to increasingly sophisticated musical robots (see \cite{bretan2016survey, weinberg2020robotic} for an overview). Existing systems span percussion robots such as Cog \cite{williamson1998rhythmic} and Shimon \cite{weinberg2007design}, string robots including Violin MUBOT \cite{kajitani1999development} and the Toyota Violin Robot, and wind robots such as the Waseda Flute and Saxophone Robots \cite{solis2009development, solis2010development, lin2019development, jia2019design} and Ob \cite{maes2011man}. These systems replicate key aspects of instrument sound production and have been primarily developed for performance or for studying human–robot interaction. As a result, their potential as experimental tools for investigating fundamental questions in music cognition remains underexplored.

In this position paper, we propose a novel use of musical humanoid robots as experimental interfaces for hypothesis-driven research in music-evoked emotions. Unlike prior work in robotic musicianship that focuses on performance or interaction design, this work positions humanoid robots as platforms that translate abstract musical structures and expressive intent into physically instantiated, controllable actions. This perspective enables the systematic manipulation of embodied and interactive aspects of musical performance that are otherwise difficult to isolate in conventional paradigms.

This paper makes three contributions:
\begin{enumerate}[noitemsep,nolistsep,label=(\roman*)]
    \item we introduce humanoid musical robots as experimental interfaces for investigating music-evoked emotions
    \item we present a case study of the anthropomorphic saxophonist robot WAS-5 as technical feasibility evidence, demonstrating reproducible acoustic control and closed-loop interaction capabilities required for future music-emotion experiments
    \item we present a set of theoretically grounded experimental paradigms that leverage capabilities afforded by musical humanoid robots.
\end{enumerate}

Our work establishes the technical feasibility of musical humanoid robots as experimental interfaces capable of supporting future controlled investigations of music-evoked emotion.

\section{Robots as Experimental Interfaces}
Current research on music-evoked emotions has largely relied on preselected, pre-recorded auditory stimuli, delivered over speakers or headphones and presented under controlled laboratory conditions \cite{warrenburg2020choosing, trost2024live}. While this approach has yielded important insights into perceptual and cognitive mechanisms underlying music-evoked emotions, it implicitly assumes that auditory information alone is sufficient in eliciting affective responses during music perception. However, growing evidence suggests that music perception is inherently multimodal, involving not only sound but also visual \cite{tsay2013sight}, somatosensory \cite{cameron2022undetectable}, and social \cite{eerola2018music} cues. As a result, current paradigms are limited by their ability to systematically manipulate embodied and interactive aspects of musical performance while maintaining experimental control.

We propose that humanoid musical robots should not only be understood as performance systems, but also as experimental interfaces for studying music-evoked emotion. By \textit{experimental interface}, we refer to a setup that allows the observation of changes in user behaviour or responses following the controlled manipulation of experimental variables \cite{mackenzie2013designing}. In this view, humanoid robots translate musical expressive parameters into physically instantiated, controllable actions that can be systematically manipulated. Notably, the role of the robot is not to replicate human performance \textit{per se}, but to provide a structured and reproducible medium through which the contribution of musical variables on emotional responses can be isolated. 

Humanoid robots offer three key properties that make them well-suited for this role. First, they enable parametric control over embodied aspects of performance, such as movement dynamics, gesture timing, and expressive cues, while maintaining precise control over acoustic output. Second, they provide reproducibility, allowing performances to be repeated across trials with minimal variability, which is difficult to achieve with human performers. Third, they support the decoupling and recombination of modalities, which enable the independent manipulation of multimodal and interaction-related variables. Together, these properties allow researchers to clarify underlying mechanisms and move towards controlled investigation of how different components of musical performance contribute to emotional experience.

Importantly, we do not claim that all musical experiences are emotional, nor that humanoid robots can fully replicate the richness of human performance. Rather, we propose that in contexts where affective responses are of interest, robots provide a complementary methodological tool that enables controlled investigation of mechanisms that are otherwise difficult to isolate. This complements, rather than replaces, studies using human musicians. Importantly, humanoid robots are not intended to replicate the full richness or ecological validity of live human performances. Rather, their value lies in providing experimentally controllable and reproducible embodiments that complement studies using human musicians, particularly when isolating specific multimodal mechanisms of music-evoked emotion.

In the following sections, we ground our position with respect to our current understanding of music-evoked emotions, present a case study of a humanoid musical robot, and outline research directions that become accessible under this framework.

\section{Cognitive Mechanisms of Music-Evoked Emotions}
\subsection{From perception to evoked emotions}\label{subsec:perception}
Existing research suggests that incoming sensory information from music is actively abstracted by the brain via two distinct systems of musical representations: tonal and metrical \cite{mehr2025core}. Tonal and metrical perception depend on sensory-acoustic constraints and statistical regularities learnt from extended exposure to music of a given style \cite{cheung2024cognitive, mehr2019universality}. The tonal system processes pitches not in isolation, but based on grouping mechanisms in terms of proximity of their relative pitches, as well as tonal hierarchies, where some pitches in a musical scale are considered more important with respect to the tonal centre established by its preceding context \cite{mehr2025core, rohrmeier2012predictive}. Similarly, beats belong to a metrical hierarchy, such that beats placed or omitted at certain locations in a meter (recurrent patterns of pulses) are considered more important than others \cite{vuust2014rhythmic}.

While the interaction of tonal and metrical representations engender expectations that are confirmed or violated, which in turn elicit an emotional response \cite{salimpoor2015predictions, cheung2022separating}, how this is combined with other musical information such as timbre, semantic information from lyrics, and multimodal associations such as body movements, remain unclear \cite{cheung2019uncertainty}. This calls for experimental paradigms where such elements can be systematically dissociated.

\subsection{Models of music-evoked emotions}\label{subsec:emotions}
Current models of music-evoked emotions characterise affective responses elicited by music in different levels of abstraction \cite{eerola2018music, eerola2012review}: At the most rudimentary level, affective responses are described in terms of core affects (valence and arousal) that can be represented by a two-dimensional circumplex model. A higher level description uses categorical emotions (e.g., happy, sad) that are \textit{sensed} and often regarded as basic. At the top of the hierarchy, elicited responses are explained in terms of complex emotions that are \textit{experienced} and described using abstract concepts such as wonder and transcendence. Music-evoked emotions thus emerge from the interaction between perceptual, cognitive, and embodied mechanisms.

One influential model that explains the mechanisms through which music can induce emotions is the BRECVEMA framework \cite{juslin2013everyday, juslin2008emotional} (see \cite{eerola2012review} for a review of other models). These include lower-level \textit{mapping} mechanisms that directly relate sound to affect, such as brain stem reflex, expectancy, entrainment, as well as embodied and socially grounded processes such as visual imagery and emotional contagion---the internal mirroring of emotions in the audience as expressed by the performer. Higher-level \textit{evaluative} mechanisms further involve evaluative conditioning, episodic memory, and aesthetic appraisal. 

That these mechanisms require the integration of auditory, visual, motor, and social information speaks for the need of a multimodal and embodied approach to investigate music-evoked emotions. By enabling reproducible, dissociable, and controlled manipulation of movement dynamics, gesture, and perceived agency alongside acoustic output, embodied music systems such as humanoid musical robots allow specific components of emotion induction (e.g., entrainment, contagion, and social interaction) to be systematically investigated.



\section{What Humanoid Robots Enable}\label{sec:robots}
Here, we explain how musical humanoid robots can serve as an invaluable experimental interface towards understanding music-evoked emotions, particularly in three directions: (i) the multimodal nature of music perception, (ii) the limits of emotion contagion, and (iii) the role of social dynamics. 

\subsection{Multimodal dissociation}\label{sec:robots_i}
While existing knowledge on music-evoked emotions was built on music as audio alone, increasing studies show that our musical experience is multimodal. For example, music presented in both auditory and visual modalities were rated as having a higher aesthetical experience and show different physiological responses compared to auditory alone \cite{czepiel2023aesthetic}, and participants appear to evaluate musical performances primarily using visual instead of auditory information \cite{tsay2013sight}. Furthermore, motor synchrony to dance music has shown to be related to imperceptible very low-frequency sounds at 8-37 Hz \cite{cameron2022undetectable}. Listeners also showed enhanced activation of the mesolimbic reward network to live music compared to pre-recorded music \cite{trost2024live}. Central to these studies is the claim that the sound produced via the physical interaction of acoustic, vibrotactile, and visual cues in music performed using actual instruments cannot be adequately reproduced by speakers alone. Although live music performance in itself does not necessitate humanoid robots, that the performance can be exactly controlled and replicable alleviates confounds that arise from variability in each human performance. That is, musical humanoid robots enable the dissociation of sound production and other emotional cues in a parametrically controlled manner.

\subsection{Controlled emotion contagion}\label{sec:robots_ii}
Another direction is to describe the precise boundary and parameters through which musical expression is transferable to the listener. Given that emotion contagion involves the internal mimicry of the performers' emotional expression, then what is the degree of anthropomorphisation (similarity to a human) required for this mimicry to occur? Current generative models of music assume that musical expression is encoded in terms of slight deviations in note onsets, velocities, and timbre \cite{cancino2018computational}, suggesting that none at all is required. This implies that it is only the \textit{belief} of the performers' emotional expression that matters in inferring intentionality, for which the physical robot acts as a strong, modifiable prior. 

Furthermore, humanoid robots allow us to clarify which actions are relevant to expression. This is important as recent research show that pianists' expressive touch is demonstrated via differences in piano key motion \cite{kuromiya2025motor}. Humanoid robots offer a controllable framework to categorise and generalise these findings. Likewise, conductors' hand movements contain gesture information that signals differences in musical expression \cite{luck2010perception}. This can be explicitly tested by systematically examining actions that convey such expressive information. Examples of such controllable parameters include smoothness/jerkiness of movements, joint velocity, head-orientation, gaze direction and latency, breathing motion, degrees of freedom in a limb, as well as microtiming deviations.

\subsection{Social dynamics and co-regulation}\label{sec:robots_iii}
While the social function of music is apparent, how social dynamics modify music-evoked emotions is not well understood. Nevertheless, early studies suggest that synchronised tapping to the beat is related to increased trust and cooperation in adults and children \cite{koelsch2014brain}, as well as differential responses when listening in a group versus alone \cite{eerola2018music}. Musical humanoid robots can fill in this gap by enabling a social environment with multiple robots actively performing or reacting to music with the listener in a controllable, co-creative set-up. Degrees of anthropomorphisation and mimicry to listeners' behaviour could further be tunable parameters that test their relevance on the social aspects of music-evoked emotions. In other words, synchrony between robot and human could be leveraged as a variable for real-time adaptation in musical co-creation paradigms.

\section{Case Study: WAS-5 as Technical Feasibility Evidence}
\begin{figure}[t]
		\centering
		\includegraphics[width=0.5\columnwidth]{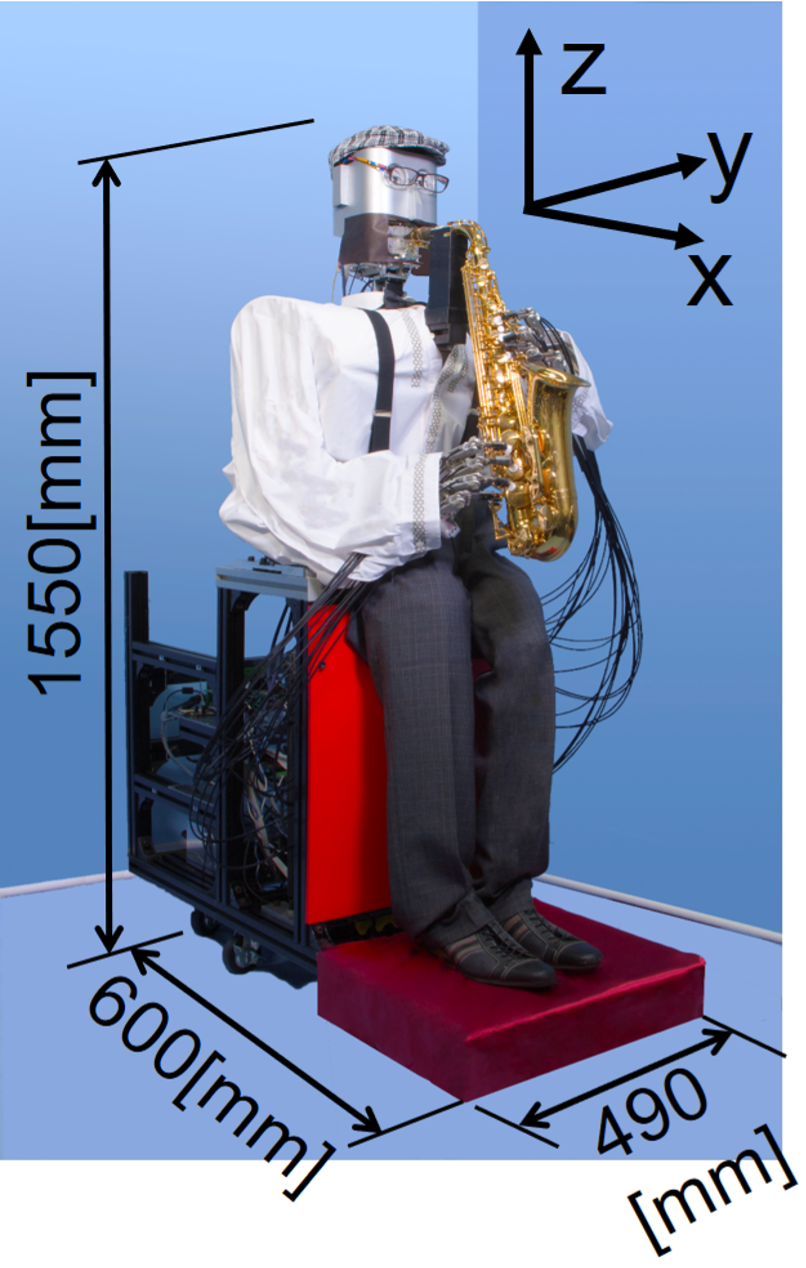}
		\caption{The Waseda Anthropomorphic Saxophonist (WAS-5) musical humanoid robot \cite{lin2019development, jia2019design}.\label{fig:WAS-5}}
\end{figure}

In this section, we introduce the Waseda Anthropomorphic Saxophonist (WAS-5) robot (see Figure \ref{fig:WAS-5}) as a case study to show its feasibility in enabling parametric control over acoustic and interaction-level variables relevant to music-evoked emotion research. WAS-5 is the fifth generation of the Waseda Anthropomorphic Saxophonist robot \cite{solis2010development, lin2019development, jia2019design}, a humanoid robot designed to play a standard alto saxophone by mimicking human respiratory airflow mechanisms, as well as finger, eye, eyelid, and eyebrow movements with 31 degrees of freedom. 

\subsection{Advanced lip mechanism for fine-grained sound pressure control}
To simulate the human respiratory tract, WAS-5 utilises an air pump and proportional valves to deliver precise airflow to the mouthpiece. In addition, its tongue and lips are made from a styrene thermoplastic elastomer to control airflow within its oral cavity with high biomimetic accuracy.

\begin{figure}[t]
		\centering
		\includegraphics[width=0.6\columnwidth]{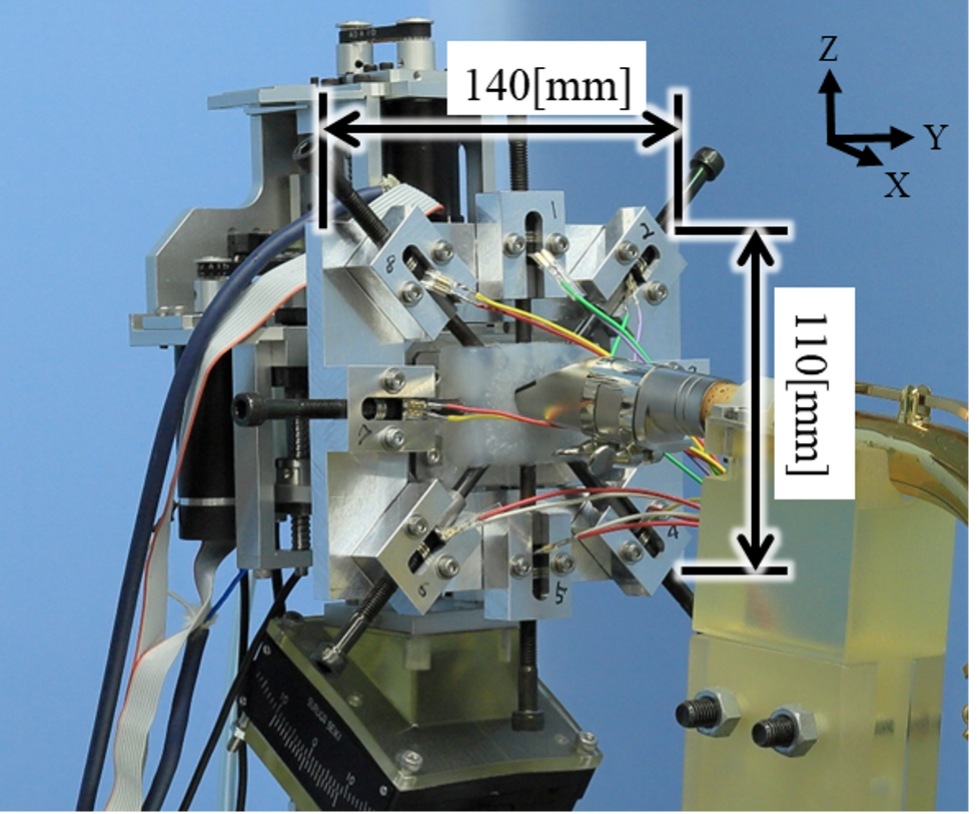}
		\caption{The 8-directional lip mechanism in WAS-5 for precise radial airflow control.\label{fig:WAS-5 lip}}
\end{figure}

In human performance, the embouchure involves coordinated muscle activation around the lips that enable the performer to apply distributed and directionally controlled forces to the reed. A new lip mechanism implemented on WAS-5 employs an 8-directional actuation structure to better approximate the circumferential contact and radial force application observed in human embouchure formation (see Figure \ref{fig:WAS-5 lip}). 

Measurements show that this 8-directional lip mechanism increased the achievable sound pressure range by up to 33.8\% compared to its previous configuration that was limited to two opposing directions, reaching a dynamic range of 24.3 dB when sounding pitch E4 and 26.5 dB when sounding pitch F4. This stable yet flexible interface can therefore support fine and reproducible control over vibration characteristics and airflow, enabling the robot to sustain the wide dynamic range essential for emotion expression.

From the perspective of music-evoked emotion research, these results provide technical feasibility evidence that the robot can reproducibly manipulate embodied acoustic parameters relevant to future music-emotion studies. Previous work has suggested that physical parameters of sound production such as embouchure adjustment, mouthpiece bite force, and reed depth contribute to acoustic features associated with emotion expression \cite{almeida2023expressive}. The WAS-5 robot thus provides the capability required for future parametric investigations of how embodied acoustic features influence music-evoked emotions via explicit mechanical control.

\subsection{Robot-Assisted Music Interaction}
While precise control of sound production is crucial, investigating music-evoked emotions also requires understanding how listeners and performers engage in dynamic contexts. To this end, WAS-5 is also capable of interacting with the listener in a closed-loop design, enabling real-time coupling between musical output and human behaviour. 

In one experiment, participants were asked to tap their feet to a simple six-note melody as performed by WAS-5 under three experimental conditions. In the Control condition, the robot performed the melody at fixed tempi of 60 bpm and 120 bpm in 60 s intervals. In the Follower condition, the robot assumed a passive role and continuously adjusted its tempo according to participants' self-designated fast and slow tapping rates. In the Leader condition, the robot assumed an active role and modulated its tempo to guide the participant towards tapping at the predefined target tempi (60 or 120 bpm). In both adaptive conditions, tempo modulation was implemented via real-time tempo estimation based on video tracking of participants' movements. 

\begin{figure}[t]
		\centering
		\includegraphics[width=0.9\columnwidth]{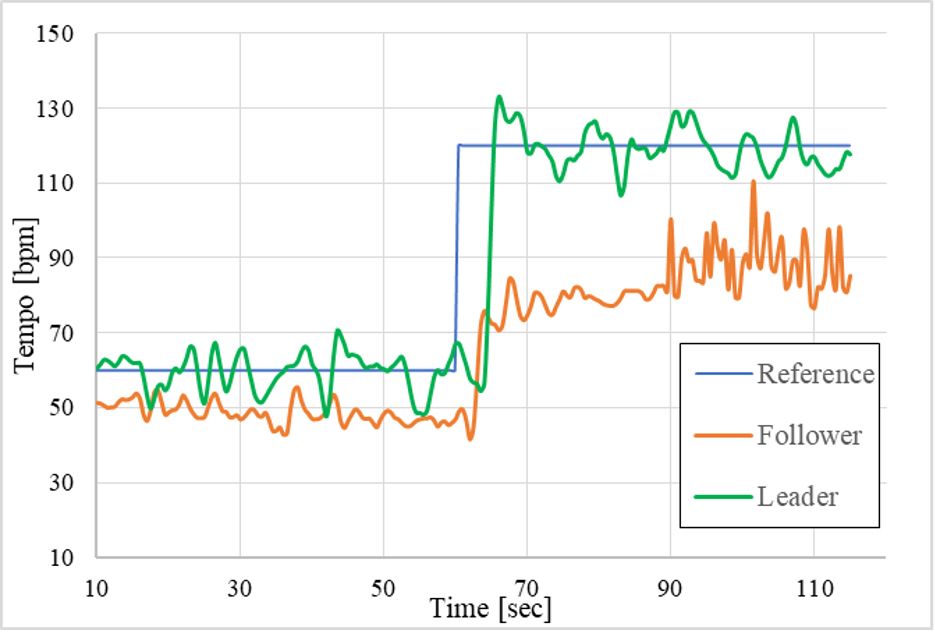}
		\caption{Feet-tapping dynamics of a representative subject, guided by WAS-5 in the Leader condition and followed by the robot in the Follower condition.\label{fig:WAS-5 interaction}}
\end{figure}

Results (see Figure \ref{fig:WAS-5 interaction}) showed that WAS-5 could reliably adapt its musical output to participants' behaviour in the follower condition, and systematically influence participants' tapping behaviour in the leader condition. These findings demonstrate that WAS-5 can reliably manipulate agency and interpersonal synchronization under controlled conditions. Importantly, this experiment evaluates behavioral synchronization rather than music-evoked emotion itself. Instead, it provides technical feasibility evidence that the platform supports closed-loop interaction paradigms which can subsequently be extended with subjective emotion ratings and physiological measures to investigate music-evoked emotional responses directly.

Together, these results show WAS-5 as an example humanoid musical robot that provides (i) precise acoustic control, (ii) reproducible performance, and (iii) closed-loop interaction capabilities, which are key to experimentally manipulating multimodal and social variables in music-evoked emotion research.

\section{Experimental Paradigms Enabled by the System}\label{sec:experiments}

In this section, we present paradigms as experimentally testable research directions. Their purpose is to illustrate how the demonstrated capabilities of musical humanoid robots may be leveraged to investigate specific mechanisms underlying music-evoked emotion in future empirical work. Building on the capabilities demonstrated above, we now outline some concrete experimental paradigms that operationalise the dimensions of multimodal integration, emotion contagion, and social interaction for research in music-evoked emotions. In each case, the use of robots is crucial as we require exact reproduction of a musical performance, direct manipulation of parameters relating to anthropomorphism, and precise movement synchronisation across multiple agents. We also summarise the mechanisms and main dependent variables explored in each paradigm in Table \ref{tab:paradigms}.

\begin{table*}[t]
\centering
\caption{Summary of proposed paradigms enabled by humanoid robots for music emotion research, as well the underlying mechanisms targeted, and main psychological outcome variables that could be measured.}
\label{tab:paradigms}
\begin{tabular}{p{0.22\textwidth} p{0.28\textwidth} p{0.40\textwidth}}
\\
\hline
\textbf{Paradigm} & \textbf{Target mechanism} & \textbf{Main dependent variables} \\
\hline
Embodiment &
Multimodal integration &
Engagement, enjoyment, trust \\

Emotion contagion &
Emotional mirroring &
Perceived emotion, intentionality \\

Social co-regulation &
Synchrony &
Trust, engagement, arousal \\
\hline
\end{tabular}
\end{table*}

\subsection{Embodiment and emotions}
Here we test the effects of different levels of robotic embodiment on music-evoked emotions. The participant is seated behind a veil for visual isolation. We first compare live music performed by a musical humanoid robot versus the same performance as presented via speakers, matched as closely in loudness and proximity as possible. Next, we repeat the same presentation with the veil removed, except there will now be two presentations via speakers, with the robot either remaining static or moving synchronously to the music. After each presentation, we measure the participants' engagement, enjoyment, and trust in the robot. We hypothesise that participants will give a higher engagement and enjoyment rating to music performed live by the robot in both veiled and unveiled conditions compared to music reproduced via speakers, with the highest scores in the former. Findings from this experiment will enable a deeper understanding on the role of embodiment on music interaction towards a richer musical cocreation experience between human and robots.

\subsection{Controlled emotion contagion}
This experiment examines the extent visual cues signalled by the robot influences emotion contagion. The participant observes multiple repetitions of the same live music performance by a humanoid robot. At each repetition, features related to robot anthropomorphism, such as body sway, eyebrow/lid movement, facial expression, or breathing-related torso movement, are parametrically varied at predefined step sizes (i.e., constant auditory but varied visual output). After each presentation, the participant provides ratings on the perceived and felt intensity and intentionality of the emotions expressed. Given the importance of body sway in expressive music performance and rhythmic entrainment \cite{chang2019body}, we expect this feature to be most relevant in emotional expression. Results allow for the identification and quantification of visual features most salient for conveying music emotions and could be incorporated in future musical humanoid robot designs. 

\subsection{Social co-regulation}
In this experiment, the participant is seated in between two humanoid robots in the audience. On stage is another musical humanoid robot performing live. At each repetition, the audience robots sway together in phase with the performer robot, together out of phase with the performer, or out of phase with each other and the performer. We measure participants' own body movement and electrodermal activity during the music, as well as collect behavioural ratings on their engagement and arousal, and ratings on how much they trust their neighbouring robots. Another variant would be to also vary the number of robots in the audience in a scaled paradigm. We hypothesise that synchronised body sway with the most audience robots would result in the highest engagement, arousal, and trust in the neighbouring robot, as well as synchronous movements by the participant. Findings from this experiment could inform the effects of social entrainment on music-evoked emotions and contribute towards developing more engaging creative systems.

\section{Limitations}\label{sec:limitations}
Our proposal on using musical humanoid robots as experimental interfaces to explore music-evoked emotions is not without limitations. First, the viable experimental state-space is constrained by current technological capabilities of a robot. For example, limitations in materials, sensors, and actuators could limit the performance of the robot and consequently restrict the questions that could be realistically explored. Nevertheless, although the paradigms presented previously were conceptual, they lie within the capabilities of existing musical humanoid robots. Second, constructing and calibrating the robot may be a challenge in itself and could introduce additional technical and logistical hurdles that must be overcome beforehand \cite{bretan2016survey}. A trade-off between the benefits of building physical humanoid robots versus the design freedoms offered by virtual reality should thus be considered (see \cite{pessanha2021virtual} for a relevant discussion). Third, the receptivity and attitude towards robots are individually and culturally dependent \cite{chien2018effect}. This means that results only reveal contemporary attitudes and may not generalise across all listeners geographically and over time. A multicultural approach with the potential to study their effects longitudinally may be required. Fourth, the musical humanoid robots should be designed to avoid the uncanny valley effect induced by perceptual mismatch arising from them having an almost-human resemblance. Fifth, despite the lack of physical grounding, virtual musician robots may also offer additional flexibility in research design  \cite{pessanha2021virtual}. Finally, the present work demonstrates only technical feasibility rather than empirical evidence that humanoid robots modulate music-evoked emotion. Establishing such evidence will require future experiments incorporating validated subjective emotion questionnaires together with physiological and behavioural measures.

\section{Conclusion}\label{sec:conclusion}
In this position paper, we introduced musical humanoid robots as experimental interfaces for investigating music-evoked emotions. We demonstrated through a case study of the WAS-5 robot that such systems can achieve reproducible control of embodied acoustic parameters and closed-loop interaction with human behaviour, allowing humanoid robots to systematically manipulate multimodal and social aspects of musical performance. We further outlined experimentally testable paradigms targeting embodiment, emotion contagion, and social co-regulation. Findings from these research directions will prove invaluable for designing future creative systems with deepened understanding of agency and transmission of expressive intentionality, as well as further knowledge on how music is often regarded one of life's most important pleasures.

\begin{acknowledgments}
This work was supported by Sony Computer Science Laboratories and JSPS Grant-in-Aid for Early-Career Scientists [23K12755]. The authors would also like to thank Tokyo Women’s Medical University/ Waseda University Joint Institution for Advanced Biomedical Sciences (TWIns) and Humanoid Robotics Institute (HRI).
\end{acknowledgments} 
	
	\bibliography{mainbib}
	
\end{document}